\documentclass[runningheads]{llncs}
\usepackage[T1]{fontenc}
\usepackage{graphicx}
\usepackage{amsmath}
\usepackage{url}
\usepackage{tikz}
\usetikzlibrary{arrows.meta,positioning,fit,shapes.misc, calc}
\definecolor{pearlgreen}{RGB}{206,230,206}
\definecolor{usersimblue}{RGB}{202,223,248}
\definecolor{recsimpurple}{RGB}{228,214,240}
\definecolor{maskyellow}{RGB}{255,230,150}

\usepackage{xcolor}
\usepackage[numbers,sort&compress]{natbib}

\begin{document}
\title{\texttt{Interplay}: Training Independent Simulators for Reference-Free Conversational Recommendation}
\titlerunning{\texttt{Interplay}: Training Independent Simulators for Reference-Free CRS}
%
\authorrunning{Ramos et al.}
%
\author{Jerome Ramos\inst{1} \and
Feng Xia\inst{2} \and
Xi Wang\inst{2} \and
Shubham Chatterjee\inst{3} \and
Xiao Fu\inst{1} \and
Hossein A. Rahmani\inst{1} \and
Aldo Lipani\inst{1}
}

\institute{
University College London, London, UK\\
\email{\{jerome.ramos.20,xiao.fu.20,hossein.rahmani.22,aldo.lipani\}@ucl.ac.uk}
\and
University of Sheffield, Sheffield, UK\\
\email{\{fxia8,xi.wang\}@sheffield.ac.uk}
\and
Missouri University of Science and Technology, Rolla, USA\\
\email{shubham.chatterjee@mst.edu}
}

\maketitle              
\begin{abstract}
Training conversational recommender systems (CRS) requires extensive dialogue data, which is challenging to collect at scale. To address this, researchers have used simulated user-recommender conversations. Traditional simulation approaches often utilize a single large language model (LLM) that generates entire conversations with prior knowledge of the target items, leading to scripted and artificial dialogues. We propose a reference-free simulation framework that trains two independent LLMs, one as the user and one as the conversational recommender. These models interact in real-time without access to predetermined target items, but preference summaries and target attributes, enabling the recommender to genuinely infer user preferences through dialogue. This approach produces more realistic and diverse conversations that closely mirror authentic human-AI interactions. Our reference-free simulators match or exceed existing methods in quality, while offering a scalable solution for generating high-quality conversational recommendation data without constraining conversations to pre-defined target items. We conduct both quantitative and human evaluations to confirm the effectiveness of our reference-free approach.

\end{abstract}
\section{Introduction}
\label{sec:introduction}
Conversational Recommender Systems (CRS) are mixed-initiative dialogue systems that offer personalized, context-sensitive recommendations through natural language conversations \cite{sun2018conversational}. Unlike traditional systems that rely on static user-item interactions, CRS allow dynamic, interactive feedback from users and enable both parties to guide the conversation.

The development of CRS depends on rich conversational data, but collecting human annotated conversations is costly and limited in quality. Previous work highlights key issues in crowd-sourced CRS datasets: (1) \textbf{Generic utterances}: Due to lack of genuine preferences, workers often say vague statements like ``I like most genres,'' limiting personalization~\cite{kim-etal-2024-pearl}; (2) \textbf{Lack of context}: ~\citet{qiu2024unveiling} showed that crowd-sourced conversations often lack depth and context, making it difficult for CRS to accurately model user preferences and provide relevant suggestions; and (3) \textbf{Limited domain expertise}: ~\citet{qiu2024unveiling} noted that crowd-workers often lack domain expertise, resulting in weak recommendations and poor explanations. ~\citet{joko2024laps} further showed that even well-designed crowd data lacks multi-session realism and is hard to scale without sacrificing authenticity. Human-written conversations are also costly and labor-intensive to collect at scale.

The growing popularity of large language models (LLMs) has led to the use of LLM simulators to generate authentic and context-rich conversations that are critical for training CRS. ~\citet{ZhouImprovingCRS2020} found that topic-guided dialogues improve recommendations, underscoring the need for synthetic data that mimic natural interactions. Recent approaches use LLMs like GPT-4o to generate CRS conversations~\cite{zhu2024reliablesimulatoranalysislimitations, kim-etal-2024-pearl, liang-etal-2024-llm}, but generally rely on reference-dependent generation, feeding target items to simulators in advance. This causes: (1) data leakage (the simulators knows the answer beforehand) and (2) rigid dialogues (exploration is limited, reducing realism). Additionally, using general-purpose LLMs as static role-players for both conversational parties is problematic~\cite{kongPlatoLMTeachingLLMs2024}. These models are designed as helpful assistants rather than realistic user simulators with diverse and sometimes inconsistent preferences. The resulting conversations often lack the natural exploration, uncertainty, and flexibility that characterize genuine human-recommender interactions.

\begin{figure}
\centering
\includegraphics[width=0.9\linewidth]{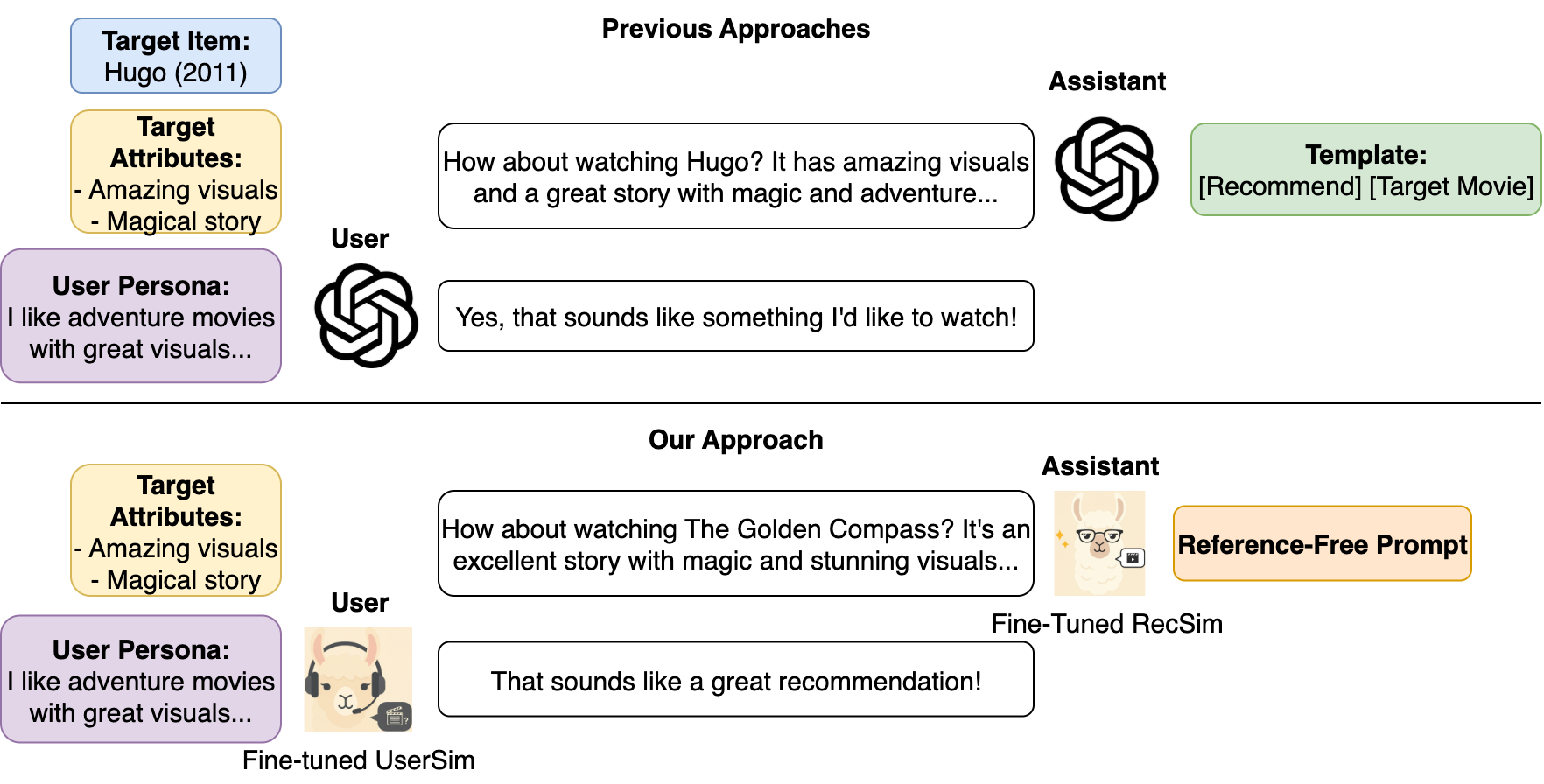}
\caption{Unlike prior work that uses templated dialogues and target-aware static LLMs, our method independently trains user and recommender simulators without predetermined items or actions, enabling diverse, realistic interactions.}
\label{fig:intro-figure}
\end{figure}

In this chapter, we address these limitations through an interplay approach that trains two specialized models to interact independently. As depicted in Figure \ref{fig:intro-figure}, our method fine-tunes separate user and recommender simulators on existing CRS data using a reference-free framework. Crucially, neither simulator has access to target items during generation. The user simulator operates with preference summaries and attribute descriptions, providing realistic feedback without knowing specific targets. The recommender simulator generates contextually appropriate recommendations based solely on the evolving conversation. This design enables genuine exploration where recommendations emerge naturally from dialogue rather than predetermined scripts.

Our approach offers several advantages over existing methods. The reference-free design eliminates data leakage by ensuring that simulators discover information through conversation rather than having it predetermined for them. Independent training of specialized models creates more realistic user and recommender behaviors than using general-purpose LLMs for both roles. The interactive nature allows flexible conversations that can naturally evolve in multiple directions, better reflecting the complexity of real recommendation scenarios.
Through comprehensive evaluations, we demonstrate that our simulators generate conversations comparable in quality to existing approaches while addressing the methodological limitations that have constrained previous work. The resulting synthetic dialogues exhibit natural exploration patterns, genuine uncertainty, and flexibility that characterize human conversations.

We make the following contributions in this chapter:

\begin{enumerate}
    \item We propose a reference-free simulation framework that removes the oracle problem by training independent user and recommender simulators without access to the target item, allowing for the exploration of realistic preferences and flexible recommendation behavior.
    \item  We show that simulators trained on target attributes, not items, outperform larger general models: Our 8B recommender achieves 15× higher accuracy than 32B baselines while preserving strong response diversity.
    \item Through a qualitative user study, we show that independent simulators match reference-dependent methods in dialogue quality across six metrics, while using smaller open-source models for a more scalable and efficient solution.

\end{enumerate}

\section{Related Works}
\label{sec:related_works}
\noindent \textbf{Conversational Recommender Systems.}
CRS integrate dialogue interfaces with personalized recommendation engines, allowing users to request and receive item suggestions through natural language interaction~\cite{10.1145/3453154,sun2018conversational}. Unlike one-shot recommenders, CRS conduct multi-turn conversations to elicit user preferences, clarify requirements, and offer explanations~\cite{10.1145/3453154}. Recent advances have shifted from modular pipelines~\cite{kang-etal-2019-recommendation} to end-to-end models~\cite{li2023conversation} and prompt-based approaches that achieve strong zero-shot performance~\cite{gao2023chat,10.1145/3627043.3659574}. However, these systems face challenges in preference modeling, with methods such as CORAL that use LLMs to handle contrasting user preferences~\cite{kook-etal-2025-empowering}.

\noindent \textbf{User Simulators.}
User simulators enable scalable CRS training and evaluation without human participants~\cite{10.1145/3394486.3403202}. Early simulator techniques were rule-based but too rigid~\cite{kang-etal-2019-recommendation}, while recent approaches leverage LLMs with persona descriptions for more realistic responses. However, a critical limitation is reference dependency: many methods provide simulators with target items, creating artificial constraints~\cite{wang-etal-2023-rethinking-evaluation}. Frameworks like CSHI balance realism and consistency through human-in-the-loop guidance~\cite{10.1145/3696410.3714858}. Our work addresses these issues through independent simulator training.

\noindent \textbf{Dialogue Data Collection and Generation.}
CRS research faces a trade-off between data quality and scalability. Human-centric methods produce high quality but are expensive~\cite{10.1145/3394486.3403202}, while crowdsourcing offers scale but suffers from generic responses and quality issues~\cite{kim-etal-2024-pearl}. LLM-based generation has emerged as a solution, using techniques such as template-driven approaches~\cite{liang-etal-2024-llm} and active synthesis~\cite{surana2025reviews}. However, maintaining reference-independence while ensuring diversity remains challenging.

\noindent \textbf{Interactive Simulation and Evaluation Challenges.}  
Recent multi-agent approaches in CRS use independent user and recommender simulators to foster natural interactions. Methods like MACRS~\cite{fang2024multiagent} and Collaborative Self-Play~\cite{eisenstein2025dont} leverage multiple LLMs with distinct roles, while reinforcement learning-based systems such as CRSIRL~\cite{chu2023multi} and AMIE~\cite{tu2024conversationaldiagnosticai} explore self-play and intrinsic reward modeling. However, true agent independence remains unexplored. Simultaneously, evaluation challenges persist: datasets like ReDial~\cite{li2018towards} and OpenDialKG~\cite{moon-etal-2019-opendialkg} suffer from data leakage, inflating performance through exposed targets~\cite{wang-etal-2023-rethinking-evaluation}. Newer methods, such as FACE~\cite{joko2025face}, aim to assess the quality of the dialogue without references, addressing the disconnect between traditional metrics and human judgment~\cite{10.1145/3394486.3403202} and highlighting the need for reference-free, target-agnostic evaluation.

\section{Methodology}
\label{sec:methodology}


We propose \texttt{Interplay} -- a framework that trains independent user and recommender simulators to interact without predetermined outcomes, allowing genuine exploration and flexible conversation dynamics that mirror real-world recommendation scenarios.

\subsection{The Core Insight}
The core challenge in simulating conversational recommendation lies in replicating the natural tension of real interactions: users bring preferences but stay open to discovering unexpected options, while recommenders must explore needs without knowing what will click. This tension is lost when a single model plays both roles with shared knowledge.  Our approach addresses this challenge through two key innovations. First, we train completely independent simulators, each optimized solely for its respective role, ensuring that natural behavioral differences emerge rather than being artificially programmed. Second, we remove reference dependence by using target attributes instead of items, prompting simulators to engage in genuine discovery over scripted responses.

\subsection{Breaking the Single-Target Constraint}
Prior user simulation methods are fundamentally rigid -- they condition responses on fixed target movies, making simulated users reject strong recommendations simply because they do not match a preset title~\cite{wang-etal-2023-rethinking-evaluation}. This creates an artificial constraint that real users never face. When someone seeks ``a romantic comedy for date night,'' they usually remain open to various films that meet those criteria rather than focusing on a single predetermined choice. Our user simulator captures natural flexibility by reasoning over target attributes instead of fixed items. Rather than knowing ``that the user wants Inception,'' the simulator is trained to understand preferences like ``a mind-bending thriller with complex plotting.'' This change allows the user simulator to accept a range of valid recommendations, freeing conversations from the constraint of converging on a single outcome. As a result, dialogues unfold more organically, driven by authentic attribute alignment, rather than scripted goals.

\subsection{Persona Architecture for Authentic User Behavior}
To produce realistic interactions, our user simulator integrates three persona components that capture diverse aspects of user context while preserving flexibility. The \emph{general preferences component} offers a natural language summary of the user's broad tastes, generated by analyzing a subset of their viewing history. This summary grounds the simulator in a consistent personality across conversations while still allowing variation in specific choices. The \emph{movie history component} includes three previously watched movies with corresponding reviews, randomly sampled from the user's profile to provide concrete examples of their tastes. These serve as conversational anchors, allowing the recommender to reference and build on specific, discussable preferences. The three random movies were originally sampled with the \texttt{PEARL} dataset and we reuse the same movies in order to maintain the same context and avoid data leakage. The \emph{target attributes component} then introduces preference cues extracted from a target movie's review, without revealing the movie title. This preserves authentic user intent while removing reference dependence: the simulator knows what qualities to look for, but not which item should satisfy them. This architecture enables our user simulator to maintain a consistent personality while remaining genuinely open to discovery, accurately mirroring how real users approach recommendation conversations with both clear preferences and openness to surprise.

\begin{figure}[t]
\centering
\footnotesize
\resizebox{0.9\linewidth}{!}{%
\begin{tikzpicture}[
  >=Latex,
  node distance=10mm and 14mm,
  box/.style={draw, rounded corners=2pt, align=left, inner sep=4pt, outer sep=0pt},
  data/.style={box, fill=pearlgreen},
  view/.style={box, text width=5.2cm, minimum height=16mm},
  model/.style={box, minimum width=2.8cm, align=center},
  lab/.style={font=\scriptsize, inner sep=1pt, text=black!60},
  mask/.style={fill=maskyellow, inner sep=1pt, rounded corners=1pt}
]

\node[data] (pearl) {\textbf{PEARL Dialogues}};

\coordinate (split) at ($(pearl.south)+(0,-6mm)$);

\node[view, below=18mm of pearl, xshift=-35mm] (uview) {%
\textbf{Masked view for UserSim}\\
\textit{User:} \texttt{<feedback>} …\\
\textit{Assistant:} \fcolorbox{black!20}{maskyellow}{\texttt{<MASK>}}
};
\node[view, below=18mm of pearl, xshift=35mm] (rview) {%
\textbf{Masked view for RecSim}\\
\textit{User:} \fcolorbox{black!20}{maskyellow}{\texttt{<MASK>}}\\
\textit{Assistant:} \texttt{<recommend>} …
};

\coordinate (tapL) at (uview.north |- split);
\coordinate (tapR) at (rview.north |- split);

\node[model, fill=usersimblue, below=10mm of uview] (usersim) {\textbf{UserSim}\\(LLM fine-tune)};
\node[model, fill=recsimpurple, below=10mm of rview] (recsim) {\textbf{RecSim}\\(LLM fine-tune)};

\draw[->, shorten >=2pt] (pearl.south) -- (split);

\draw[shorten >=2pt] (split) -- (tapL);
\draw[shorten >=2pt] (split) -- (tapR);

\draw[->, shorten <=2pt, shorten >=2pt] (tapL) -- (uview.north);
\node[lab, anchor=east, xshift=-3pt] at ($(tapL)!0.25!(uview.north)$) {mask assistant};

\draw[->, shorten <=2pt, shorten >=2pt] (tapR) -- (rview.north);
\node[lab, anchor=west, xshift=3pt] at ($(tapR)!0.25!(rview.north)$) {mask user};

\draw[->, shorten <=2pt, shorten >=2pt] (uview.south) -- (usersim.north);
\node[lab, anchor=east, xshift=-3pt] at ($(uview.south)!0.6!(usersim.north)$) {train};

\draw[->, shorten <=2pt, shorten >=2pt] (rview.south) -- (recsim.north);
\node[lab, anchor=west, xshift=3pt] at ($(rview.south)!0.6!(recsim.north)$) {train};

\end{tikzpicture}
}
\caption{A single PEARL dialogue is annotated with structural tokens to train two separate simulators: \emph{(left)} masks assistant turns and keeps user turns to train \textbf{UserSim}; \emph{(right)} masks user turns and keeps assistant turns to train \textbf{RecSim}.}
\label{fig:masked-views-sims}
\end{figure}
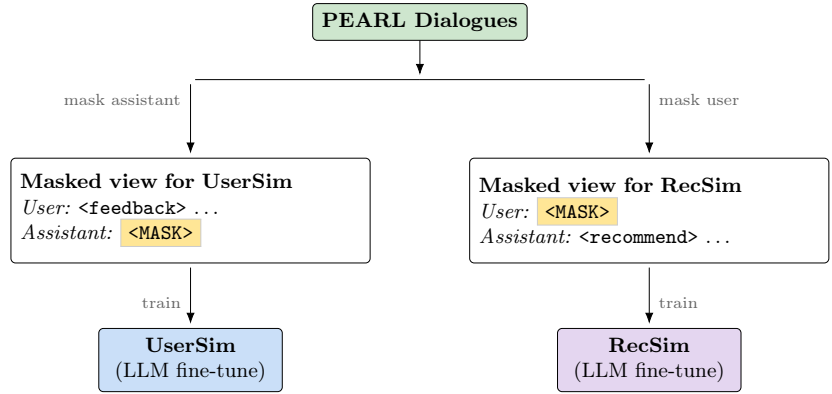

\subsection{Recommender Without Oracle Knowledge}
Our recommender simulator learns to handle preference elicitation and recommendation generation through fluid conversations, adapting dynamically to evolving user needs rather than following rigid patterns. We cast the recommender task as contextual text generation, where the model learns to balance exploration (eliciting detailed preferences) and exploitation (offering confident, informed recommendations). Crucially, the recommender has no access to target items, compelling it to build genuine understanding and match user needs organically rather than guiding conversations toward predefined outcomes. Because our user simulator accepts any recommendation that satisfies its attribute criteria, conversations naturally conclude when a suitable match emerges. This avoids forced extensions seen in systems that must reach a specific target item. Instead, interactions end organically when users express satisfaction, resulting in more realistic dialogue patterns. The recommender learns to detect these signals and adapt its strategy, shaping nuanced, feedback-driven behaviors rather than following rigid scripts.

\subsection{Independent Training for Specialized Behavior}
Our novel training strategy is composed of two key components designed to foster specialized agent behavior. First, we enforce a structured output format to make conversational actions explicit. Second, we employ a role-specific loss masking technique to ensure each simulator learns exclusively from its own turns.

\subsubsection{Structured Action Generation}
To enforce role-specific behavior, we structure the entire model output to separate the \textit{intended action} from the \textit{natural language response}. Each generated turn adheres to a strict format where an \texttt{<action>} block contains a specific command: \texttt{<recommend>}, \texttt{<inquire>}, \texttt{<greeting>}, \texttt{<disclose-goal>}, \texttt{<feedback>}, and \texttt{<accept>}. Afterwards, a \texttt{<response>} block provides the user-facing text, including the movie title, where applicable. These structural tokens are added to the tokenizer before training. This design provides two key advantages:

\begin{enumerate}
    \item \textbf{Structured Reasoning:} It imposes an explicit response template, teaching the model to first commit to an unambiguous action before generating its utterance. For example, when providing a recommendation, the model can respond with:\\\texttt{<action><recommend></action><response>}You may enjoy \texttt{<movie\_title>Good Will Hunting (1997)</movie\_title>}.
    
    \item \textbf{Simplified Analysis:} The structured output format enables reliable post-simulation parsing, allowing us to programmatically extract and analyze the sequence of actions and recommended items. Crucially, the user simulator is trained to output \texttt{<accept>} when it encounters a satisfactory recommendation. This provides a natural and concrete stopping signal, even if the original oracle item is never recommended. In contrast, oracle-based methods can only terminate once the ground-truth item is recommended, often leading to either excessively long conversations or manual intervention where the ground truth is injected after a fixed number of turns --- an unrealistic shortcut. Our reference-free design avoids this rigidity: conversations can conclude flexibly as soon as user satisfaction is reached. While the possibility of indefinitely extended dialogues still exists in principle, in practice this approach greatly reduces the need for manual intervention and more closely reflects the open-ended nature of real human–recommender interactions.
\end{enumerate}

\subsubsection{Role-Specific Loss Masking}
Training two separate simulators on a single conversational stream presents a challenge: each model must leverage the full dialogue for context while only learning to produce its own role-specific responses. We address this with our key innovation: a token masking strategy during loss computation. We calculate cross-entropy loss only over tokens corresponding to the simulator's active role:
\begin{equation}
    L_{\text{masked}} = -\sum_{t=1}^{T} M_t \cdot \log P(y_t | y_{<t}, x)
\end{equation}
where the mask $M_t$ is defined as:
\begin{equation}
    M_t = \begin{cases}
    1 & \text{if } t \in T_{\text{simulator}} \\
    0 & \text{otherwise}
\end{cases}
\end{equation}

This allows each model to see the full conversation for context while learning to generate only role-appropriate responses. This training approach avoids the role-swapping issues seen when a single model plays multiple roles. Instead, each simulator develops distinct conversational strategies: the user learns to express preferences naturally, while the recommender learns to generate recommendations and persuasive explanations. The result is two specialized models that understand dialogue dynamics but exhibit clearly differentiated personas.

\subsection{Emergent Realistic Dynamics}

When our independently trained simulators interact, their distinct objectives and perspectives produce authentic conversational dynamics that single-model setups fail to capture. The user simulator enters with clear yet flexible goals, expressing preferences naturally and staying open to unexpected but fitting suggestions. Meanwhile, the recommender must genuinely interpret these preferences, asking clarifying questions and generating recommendations based on the evolving dialogue, not preloaded answers. This independence is key to realism: the natural tension between user uncertainty and recommender confidence emerges organically from their interaction.

\section{Experimental Setup}
\label{sec:experimental_setup}
\subsection{Experimental Details}
We perform preprocessing steps to filter \texttt{PEARL}~\cite{kim-etal-2024-pearl}. Specifically, we removed turns with empty dialogues and removed conversations with movies that were not part of the IMDB dataset used to generate \texttt{PEARL}. We filtered 7.65\% of the original dataset, leaving our final dataset with 52.9k conversations. We fine-tune our both simulators using Llama 3.1 8B~\cite{grattafiori2024llama3herdmodels} and Qwen 3 8B~\cite{qwen3technicalreport}. Following ~\citet{kongPlatoLMTeachingLLMs2024}, we use the base model versions rather than the instruct version to avoid the issue of pre-existing system alignment.

We perform full fine-tuning on both models for 2 epochs on 4 Nvidia A100 GPUs with a batch size of 4, gradient accumulation steps of 4, and a learning rate of 2e-4. We evaluate on the filtered test set of the \texttt{PEARL} dataset, containing 2,069 conversations with seen user personas but unseen target items.
For full implementation details, see: \url{https://github.com/jeromeramos70/interplay-sim}.

\subsection{Baselines}
Although previous works use closed-sourced models GPT3.5 or GPT4 to simulate conversations, we rely only on open-source baselines for reproducibility. Namely, we use the instruction-tuned versions of Llama3.1 70B~\cite{grattafiori2024llama3herdmodels} and Qwen3 32B~\cite{qwen3technicalreport} as zero-shot user and recommender simulators. Furthermore, we fine-tune UniCRS~\cite{wang2022unicrs}, a popular CRS model, on the \texttt{PEARL} dataset as another baseline.

For our zero-shot baselines, we construct prompts that mirror the input to our fine-tuned models. For the user simulator role, the prompt includes the user's persona (general preferences, movie history with reviews) and the target attributes. For the recommender simulator role, the prompt includes the user's persona and the full conversation history. This setup ensures a fair comparison by providing the baselines with the same contextual information as our models.

\subsection{Evaluation Metrics}
We assess our simulators with automatic metrics on two fronts: (i) \emph{multi-turn simulation quality} and (ii) \emph{single-turn response quality}. Human evaluations are described in Section~\ref{subsec:human-eval}.

\subsubsection{User Simulation Quality}
Each test dialogue has a known ground-truth target item. As the recommender proposes items, the simulator either accepts or rejects them. We summarize outcomes with three categories:

\begin{itemize}
    \item \textbf{Success Rate (SR):} The simulator accepts the ground-truth item.
    \item \textbf{Early Termination (ET):} The simulator accepts some \emph{non–ground-truth} item \emph{before} the ground-truth is even proposed.
    \item \textbf{Failure Rate (FR):} The simulator rejects all recommendations, including the ground-truth item.
\end{itemize}

Higher SR and lower FR indicate that the simulator reliably follows user goals and accepts valid recommendations. 
A moderate ET rate can also be desirable: it reflects that the simulator occasionally accepts alternative items instead of always defaulting to the ground-truth recommendation, capturing a more diverse and realistic range of user preferences. 
By definition, these three rates are mutually exclusive and together account for 100\% of dialogues.

\subsubsection{Single-Turn Response Quality}
To evaluate the quality of individual user responses, we report:

\begin{itemize}
    \item \textbf{BERTScore:} We measure the semantic similarity between the generated user responses and the ground-truth responses at each turn to evaluate contextual relevance (higher is better)~\cite{bert-score}.
    \item \textbf{Dist-4:} Measures lexical diversity as the fraction of unique 4-grams in the generated responses, computed over the whole test set (higher is better).
\end{itemize}

\subsubsection{Recommender Performance}
In addition to Dist-4, we report the following metrics for the recommender:
\begin{itemize}
    \item \textbf{Recall@1:} We check if the target item in a given turn is correct because our model is trained to only output a single recommendation in a single turn (higher is better).
    \item \textbf{Match Score:} To better evaluate recommendation quality in a reference-free context, we introduce Match Score. This metric quantifies the average similarity between the recommended items and the ground-truth movie. The score is the cosine similarity between movie embeddings from a model trained on the MovieLens-32M dataset~\cite{harper2015movielens}. This model uses TinyBERT~\cite{jiao2019tinybert} to encode semantic features (e.g., title, tags, genres) and learns interaction relevance via BPR loss~\cite{rendleBPR2009} to capture user preferences (higher is better). 
\end{itemize}

\section{Results and Discussions}
\label{sec:results}

\subsection{Realistic User Simulation without Oracle Knowledge}
\subsubsection{Multi-turn user simulation}
We evaluate whether our reference-free user simulator can appropriately accept and reject recommendations in a multi-turn setting by replaying conversations from the \texttt{PEARL} test set. Specifically, the user simulator walks through the recommender’s responses at each turn and determines whether to accept or reject the item. This setup directly tests the simulator’s ability to handle ongoing dialogue without oracle knowledge. As shown in Table~\ref{tab:user_sim_dialogue_eval}, our reference-free user simulator achieves high success rates and effectively addresses the oracle problem.

Our fine-tuned models achieve success rates of 93-95\%, substantially outperforming larger zero-shot LLMs including Llama3.1 70B (36\% SR) and Qwen3 32B (77\% SR). Critically, these results show that our simulators can recognize and accept appropriate recommendations without knowing the target movie title in advance, validating that reference-free operation preserves realistic user behavior while eliminating artificial constraints.

\subsubsection{Single-turn response quality}
We also evaluate the user simulator's single-turn response quality by generating a user reply at each turn $t$ using the chat history up to that point ($h = \{u_1, a_1, \ldots a_{t-1}\}$). Table~\ref{tab:user_sim_dialogue_eval} shows our models achieve superior BertScore compared to larger baselines while maintaining good diversity. This confirms that removing target knowledge does not compromise the quality of user responses -- our simulators generate contextually appropriate utterances that match human references without requiring predetermined outcomes.



\begin{table}[!t]
\caption{Evaluations for User Simulators.}
\label{tab:user_sim_dialogue_eval}
\centering
\scalebox{1}{

\begin{tabular}{l|ccc|ccc}
\hline
& \multicolumn{3}{c|}{\textbf{Multi-Turn}} & \multicolumn{3}{c}{\textbf{Single-Turn}} \\
\cline{2-7}
\textbf{Model} & \textbf{SR} & \textbf{ET} & \textbf{FR} & \textbf{BertScore} & \textbf{Dist-4} & \textbf{\# Words} \\
\hline
UserSim-Llama 8B & 0.9265 & 0.0498 & 0.0237 & 0.9364 & 0.3023 & 31.2219 \\
UserSim-Qwen 8B & 0.9531 & 0.0232 & 0.0237 & 0.9373 & 0.3093 & 30.7637 \\
Llama3.1 70B & 0.3615 & 0.0033 & 0.6351 & 0.8870 & 0.2786 & 43.4067 \\
Qwen3 32B & 0.7772 & 0.0097 & 0.2131 & 0.8921 & 0.3409 & 31.6797 \\
\hline
\end{tabular}}
\end{table}

\subsection{High-Quality Recommendations without Oracle Knowledge}
To determine whether the model can provide high-quality recommendations without oracle knowledge, we generate recommendations from our model on the turn when the ground truth is given. To measure the recommendation performance of our model, we report the Match Score between the recommended items and the ground truth movie per dialogue round.

Table~\ref{tab:rec_sim_recommendation_eval} provides strong evidence that specialized training enables effective recommendation without oracle knowledge. Our RecSim-Qwen8B achieves the highest performance with Recall@1 of 0.0217 and Match Score of 0.9333, significantly outperforming the larger Qwen3-32B baseline and exceeding the modular UniCRS~\cite{wang2022unicrs} system. This substantial performance gap demonstrates that fine-tuning smaller models for specific conversational recommendation tasks is more effective than relying on larger general-purpose LLMs with predetermined targets.

Additionally, Table~\ref{tab:rec_sim_recommendation_eval} shows our models generate more diverse responses than larger baselines while maintaining appropriate lengths. This indicates that our approach produces contextually rich, varied recommendations without the repetitive patterns common in oracle-based systems.

\begin{table}[!t]
\caption{Evaluations of recommendation performance.}
\label{tab:rec_sim_recommendation_eval}
\centering
\scalebox{1}{
\begin{tabular}{lcc|ccc}
\hline
\textbf{Model} & \textbf{Recall@1} & \textbf{Match Score} & \textbf{Dist-4} & \textbf{\# Words}  \\
\hline   
RecSim-Llama8B & 0.0193 & 0.9323 & 0.5243 & 37.42\\
RecSim-Qwen8B & 0.0217 & 0.9333 & 0.5358 & 37.30 \\
Qwen3 32B & 0.0014 & 0.4745 & 0.4630 & 42.21\\
UniCRS & 0.0119 & 0.9129 & 0.2147 & 29.69\\

\hline
\end{tabular}}
\end{table}

\begin{figure}
   \centering
   \includegraphics[width=0.95\linewidth]{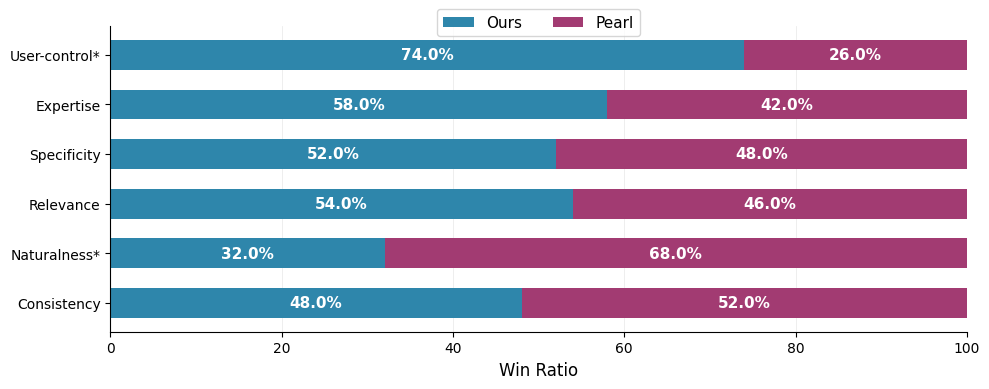}
   \caption{Human evaluation win ratio comparing reference-free dialogues to \texttt{PEARL} dataset. (* indicates p-value $<$ 0.05)}
   \label{fig:win-ratio}
\end{figure}

\subsection{Realistic Dialogue via Independent Simulation}
\label{subsec:human-eval}
To assess whether our framework generates realistic conversations, we conducted a human evaluation with 10 users across 50 dialogues pairs comparing our reference-free dialogues against \texttt{PEARL} dataset conversations using six criteria from ~\citet{kim-etal-2024-pearl}: user control, expertise, specificity of preferences, relevance, conversational flow, and consistency. Specifically, human judges were shown a head-to-head comparison between two dialogues generated using the same user profile and were instructed to select which dialogue performed better.

Figure~\ref{fig:win-ratio} reveals that our reference-free approach achieves a quality comparable to \texttt{PEARL}'s reference-dependent conversations. Evaluators found that our reference-free dialogues exhibited significantly more user control, though they did not match the naturalness of \texttt{PEARL}. This gap is partly attributable to how the models are trained: because our simulators are trained on \texttt{PEARL}’s LLM-generated text, they may inherit its stylistic patterns, so perceived naturalness likely reflects the training data’s style rather than a limitation of our reference-free design. \texttt{PEARL} was also generated using GPT-3.5, which has substantially more parameters than our 8B model, further contributing to its higher naturalness ratings. Nevertheless, our user study shows that eliminating oracle knowledge does not compromise dialogue authenticity: independent simulators produce realistic interactions through genuine exploration rather than scripted convergence on predetermined targets. Moreover, our approach achieves this quality with smaller models, significantly reducing computational costs while maintaining conversational realism.

\begin{table}[t!]
\centering
\caption{A example of the head-to-head comparison of our \texttt{Interplay} simulator (Dialogue A) versus a \texttt{PEARL} dialogue (Dialogue B). The table highlights the key turns in the conversations. Notably, \texttt{Interplay} can make high-quality recommendations without any predefined movies or target items to recommend.}
\label{tab:h2h_condensed}
\begin{tabular}{| p{0.46\columnwidth} | p{0.46\columnwidth} |}
\hline
\textbf{Dialogue A (\texttt{Interplay})} & \textbf{Dialogue B (\texttt{PEARL})} \\
\hline
\multicolumn{2}{|c|}{\textit{Initial user preference is established...}} \\
\hline
\textbf{Turn 3 - User:} I really enjoy watching movies with strong character development and intense chase/fight scenes. I also appreciate well-written scripts and great performances... & \textbf{Turn 3 - User:} I'm looking for something with action and adventure. I enjoy movies with thrilling chase and fight scenes, as well as strong performances... \\
\hline
\multicolumn{2}{|c|}{\textit{...after several exchanges and rejected recommendations...}} \\
\hline
\textbf{Turn 7 - User:} I appreciate the suggestion, but I found \textit{Superman Returns} to be lacking in intensity... the pacing felt off... I'm hoping for something with a little more edge and excitement. & \textbf{Turn 7 - User:} I appreciate the recommendation, but I'm not a big fan of \textit{Live and Let Die}... I found the villains to be a bit too over-the-top... I tend to prefer movies with a more serious tone... \\
\hline
\textbf{Turn 8 - Rec:} How about giving \textit{The Bourne Identity (2002)} a try? It has everything you're looking for - strong character development, intense chase and fight scenes, a well-written script, and excellent acting performances. & \textbf{Turn 8 - Rec:} How about watching \textit{Quantum of Solace (2008)}? It's a James Bond film that has thrilling chase and fight scenes, as well as incredibly well-written plot. \\
\hline
\textbf{Turn 9 - User:} That sounds perfect! I haven't watched \textit{The Bourne Identity} yet, so I'll definitely check it out. Thanks for the recommendation! & \textbf{Turn 9 - User:} That sounds perfect! I haven't seen \textit{Quantum of Solace} before, so I'll definitely give it a watch. Thanks for the recommendation! \\
\hline
\end{tabular}
\end{table}
\vspace{-0.8em}


\section{Limitations}
\label{sec:limitations}
\subsubsection{Domain Specificity}
Our empirical evaluation is confined to the movie recommendation domain, chosen due to the availability of large-scale, high-quality public datasets like \texttt{PEARL} \cite{kim-etal-2024-pearl}. While the \texttt{Interplay} framework is designed to be domain-agnostic, its effectiveness in other domains such as e-commerce, music, or travel has not yet been validated. Future work should focus on applying and evaluating our independent simulator approach in these different contexts, which may present unique challenges like different conversational dynamics.

\subsubsection{Constrained Conversational Scope}
The current simulation operates within a structured, task-oriented dialogue framework where conversational turns are mapped to explicit actions like \texttt{<recommend>} or \texttt{<feedback>}. This approach is beneficial for control and analysis but does not capture the full spectrum of real-world interactions. Human conversations often involve open-domain chit-chat, multi-intent utterances (e.g., asking a question and providing feedback in the same turn), and more fluid topic shifts. A key direction for future research is to enhance the simulators to handle these less structured, mixed-initiative conversations.

\subsubsection{Limitation on Recommendation Evaluation}
Although MatchScore is a useful automatic proxy for recommendation quality, it has not yet been validated as a reliable metric that aligns with real user preferences. Future work should include user studies to evaluate whether automatic metrics such as MatchScore correlates with human judgments, and more broadly evaluate whether the reference-free simulation results reflect real-world user behavior.


\section{Conclusion}
\label{sec:conclusion}

We tackle the oracle problem in conversational recommender system simulation, where access to target items creates unrealistic, constrained interactions. Our reference-free framework trains independent user and recommender simulators without predetermined outcomes, enabling more natural conversations and flexible recommendation behavior. By focusing on target attributes rather than items, user simulators accept multiple valid suggestions, while recommender simulators develop genuine preference understanding without relying on oracle knowledge.

Our approach outperforms larger models across key metrics: user simulators achieve 93–95\% success rates without target leakage, and our 8B recommenders surpass 32B baselines by 15× in accuracy while maintaining high diversity. Human evaluations confirm our dialogues match reference-dependent quality, but with far lower computational cost. This establishes a scalable, realistic simulation paradigm for CRS and paves the way for more advanced multi-agent systems in conversational AI.

\section{Acknowledgements}

This project was supported using computational resources from the Nvidia Academic Grant Program.

\section{Disclosure of Interests}
The authors have no competing interests to declare that are relevant to the content of this article.

%
%
%
\bibliographystyle{plainnat}
\bibliography{main}

@inproceedings{kongPlatoLMTeachingLLMs2024,
  title = {{{PlatoLM}}: {{Teaching LLMs}} in {{Multi-Round Dialogue}} via a {{User Simulator}}},
  shorttitle = {{{PlatoLM}}},
  booktitle = {Proceedings of the 62nd {{Annual Meeting}} of the {{Association}} for {{Computational Linguistics}} ({{Volume}} 1: {{Long Papers}})},
  author = {Kong, Chuyi and Fan, Yaxin and Wan, Xiang and Jiang, Feng and Wang, Benyou},
  editor = {Ku, Lun-Wei and Martins, Andre and Srikumar, Vivek},
  date = {2024-08},
  year = {2024},
  pages = {7841--7863},
  address = "Bangkok, Thailand",
  publisher = {Association for Computational Linguistics},
  location = {Bangkok, Thailand},
  url = {https://aclanthology.org/2024.acl-long.424},
  urldate = {2024-08-30},
  eventtitle = {{{ACL}} 2024}
}

@inproceedings{kim-etal-2024-pearl,
    title = "Pearl: A Review-driven Persona-Knowledge Grounded Conversational Recommendation Dataset",
    author = "Kim, Minjin  and
      Kim, Minju  and
      Kim, Hana  and
      Kwak, Beong-woo  and
      Kang, SeongKu  and
      Yu, Youngjae  and
      Yeo, Jinyoung  and
      Lee, Dongha",
    editor = "Ku, Lun-Wei  and
      Martins, Andre  and
      Srikumar, Vivek",
    booktitle = "Findings of the Association for Computational Linguistics: ACL 2024",
    month = aug,
    year = "2024",
    address = "Bangkok, Thailand",
    publisher = "Association for Computational Linguistics",
    url = "https://aclanthology.org/2024.findings-acl.65/",
    doi = "10.18653/v1/2024.findings-acl.65",
    pages = "1105--1120"
}

@misc{qiu2024unveiling,
      title={Unveiling User Preferences: A Knowledge Graph and LLM-Driven Approach for Conversational Recommendation}, 
      author={Zhangchi Qiu and Linhao Luo and Shirui Pan and Alan Wee-Chung Liew},
      year={2024},
      eprint={2411.14459},
      archivePrefix={arXiv},
      primaryClass={cs.CL},
      url={https://arxiv.org/abs/2411.14459}, 
}

@article{10.1145/3453154,
author = {Jannach, Dietmar and Manzoor, Ahtsham and Cai, Wanling and Chen, Li},
title = {A Survey on Conversational Recommender Systems},
year = {2021},
issue_date = {June 2022},
publisher = {Association for Computing Machinery},
address = {New York, NY, USA},
volume = {54},
number = {5},
issn = {0360-0300},
url = {https://doi.org/10.1145/3453154},
doi = {10.1145/3453154},
journal = {ACM Comput. Surv.},
month = may,
articleno = {105},
numpages = {36}
}

@inproceedings{sun2018conversational,
  title={Conversational recommender system},
  author={Sun, Yueming and Zhang, Yi},
  booktitle={The 41st international acm sigir conference on research \& development in information retrieval},
  pages={235--244},
  year={2018}
}

@inproceedings{kang-etal-2019-recommendation,
    title = "Recommendation as a Communication Game: Self-Supervised Bot-Play for Goal-oriented Dialogue",
    author = "Kang, Dongyeop  and
      Balakrishnan, Anusha  and
      Shah, Pararth  and
      Crook, Paul  and
      Boureau, Y-Lan  and
      Weston, Jason",
    editor = "Inui, Kentaro  and
      Jiang, Jing  and
      Ng, Vincent  and
      Wan, Xiaojun",
    booktitle = "Proceedings of the 2019 Conference on Empirical Methods in Natural Language Processing and the 9th International Joint Conference on Natural Language Processing (EMNLP-IJCNLP)",
    month = nov,
    year = "2019",
    address = "Hong Kong, China",
    publisher = "Association for Computational Linguistics",
    url = "https://aclanthology.org/D19-1203/",
    doi = "10.18653/v1/D19-1203",
    pages = "1951--1961"
}

@inproceedings{li2023conversation,
  title = {A Conversation is Worth A Thousand Recommendations: A Survey of Holistic Conversational Recommender Systems},
  author = {Li, Chuang and Hu, Hengchang and Zhang, Yan and Kan, Min-Yen and Li, Haizhou},
  booktitle = {Proceedings of KaRS@RecSys 2023},
  year = {2023},
  publisher = {CEUR-WS.org},
  url = {https://ceur-ws.org/Vol-3560/long1.pdf}
}

@inproceedings{10.1145/3627043.3659574,
author = {Manzoor, Ahtsham and Ziegler, Samuel C. and Garcia, Klaus Maria. Pirker and Jannach, Dietmar},
title = {ChatGPT as a Conversational Recommender System: A User-Centric Analysis},
year = {2024},
isbn = {9798400704338},
publisher = {Association for Computing Machinery},
address = {New York, NY, USA},
url = {https://doi.org/10.1145/3627043.3659574},
doi = {10.1145/3627043.3659574},
booktitle = {Proceedings of the 32nd ACM Conference on User Modeling, Adaptation and Personalization},
pages = {267–272},
numpages = {6},
location = {Cagliari, Italy},
series = {UMAP '24}
}

@article{gao2023chat,
  title={Chat-rec: Towards interactive and explainable llms-augmented recommender system},
  author={Gao, Yunfan and Sheng, Tao and Xiang, Youlin and Xiong, Yun and Wang, Haofen and Zhang, Jiawei},
  journal={arXiv preprint arXiv:2303.14524},
  year={2023}
}

@inproceedings{10.1145/3394486.3403202,
author = {Zhang, Shuo and Balog, Krisztian},
title = {Evaluating Conversational Recommender Systems via User Simulation},
year = {2020},
isbn = {9781450379984},
publisher = {Association for Computing Machinery},
address = {New York, NY, USA},
url = {https://doi.org/10.1145/3394486.3403202},
doi = {10.1145/3394486.3403202},
booktitle = {Proceedings of the 26th ACM SIGKDD International Conference on Knowledge Discovery \& Data Mining},
pages = {1512–1520},
numpages = {9},
location = {Virtual Event, CA, USA},
series = {KDD '20}
}

@inproceedings{10.1145/3696410.3714858,
author = {Zhu, Lixi and Huang, Xiaowen and Sang, Jitao},
title = {A LLM-based Controllable, Scalable, Human-Involved User Simulator Framework for Conversational Recommender Systems},
year = {2025},
isbn = {9798400712746},
publisher = {Association for Computing Machinery},
address = {New York, NY, USA},
url = {https://doi.org/10.1145/3696410.3714858},
doi = {10.1145/3696410.3714858},
booktitle = {Proceedings of the ACM on Web Conference 2025},
pages = {4653–4661},
numpages = {9},
location = {Sydney NSW, Australia},
series = {WWW '25}
}

@inproceedings{wang-etal-2023-rethinking-evaluation,
    title = "Rethinking the Evaluation for Conversational Recommendation in the Era of Large Language Models",
    author = "Wang, Xiaolei  and
      Tang, Xinyu  and
      Zhao, Xin  and
      Wang, Jingyuan  and
      Wen, Ji-Rong",
    editor = "Bouamor, Houda  and
      Pino, Juan  and
      Bali, Kalika",
    booktitle = "Proceedings of the 2023 Conference on Empirical Methods in Natural Language Processing",
    month = dec,
    year = "2023",
    address = "Singapore",
    publisher = "Association for Computational Linguistics",
    url = "https://aclanthology.org/2023.emnlp-main.621/",
    doi = "10.18653/v1/2023.emnlp-main.621",
    pages = "10052--10065"
}

@inproceedings{zhu2024reliablesimulatoranalysislimitations,
author = {Zhu, Lixi and Huang, Xiaowen and Sang, Jitao},
title = {How Reliable is Your Simulator? Analysis on the Limitations of Current LLM-based User Simulators for Conversational Recommendation},
year = {2024},
isbn = {9798400701726},
publisher = {Association for Computing Machinery},
address = {New York, NY, USA},
url = {https://doi.org/10.1145/3589335.3651955},
doi = {10.1145/3589335.3651955},
booktitle = {Companion Proceedings of the ACM Web Conference 2024},
pages = {1726–1732},
numpages = {7},
location = {Singapore, Singapore},
series = {WWW '24}
}

@inproceedings{ZhouImprovingCRS2020,
author = {Zhou, Kun and Zhao, Wayne Xin and Bian, Shuqing and Zhou, Yuanhang and Wen, Ji-Rong and Yu, Jingsong},
title = {Improving Conversational Recommender Systems via Knowledge Graph based Semantic Fusion},
year = {2020},
isbn = {9781450379984},
publisher = {Association for Computing Machinery},
address = {New York, NY, USA},
url = {https://doi.org/10.1145/3394486.3403143},
doi = {10.1145/3394486.3403143},
booktitle = {Proceedings of the 26th ACM SIGKDD International Conference on Knowledge Discovery \& Data Mining},
pages = {1006–1014},
numpages = {9},
location = {Virtual Event, CA, USA},
series = {KDD '20}
}

@misc{joko2025face,
      title={FACE: A Fine-grained Reference Free Evaluator for Conversational Recommender Systems}, 
      author={Hideaki Joko and Faegheh Hasibi},
      year={2025},
      eprint={2506.00314},
      archivePrefix={arXiv},
      primaryClass={cs.IR},
      url={https://arxiv.org/abs/2506.00314}, 
}

@inproceedings{moon-etal-2019-opendialkg,
    title = "{O}pen{D}ial{KG}: Explainable Conversational Reasoning with Attention-based Walks over Knowledge Graphs",
    author = "Moon, Seungwhan  and
      Shah, Pararth  and
      Kumar, Anuj  and
      Subba, Rajen",
    editor = "Korhonen, Anna  and
      Traum, David  and
      M{\`a}rquez, Llu{\'i}s",
    booktitle = "Proceedings of the 57th Annual Meeting of the Association for Computational Linguistics",
    month = jul,
    year = "2019",
    address = "Florence, Italy",
    publisher = "Association for Computational Linguistics",
    url = "https://aclanthology.org/P19-1081/",
    doi = "10.18653/v1/P19-1081",
    pages = "845--854"
}

@inproceedings{li2018towards,
author = {Li, Raymond and Kahou, Samira and Schulz, Hannes and Michalski, Vincent and Charlin, Laurent and Pal, Chris},
title = {Towards Deep Conversational Recommendations},
year = {2018},
publisher = {Curran Associates Inc.},
address = {Red Hook, NY, USA},
booktitle = {Proceedings of the 32nd International Conference on Neural Information Processing Systems},
pages = {9748–9758},
numpages = {11},
location = {Montr\'{e}al, Canada},
series = {NIPS'18}
}

@misc{tu2024conversationaldiagnosticai,
      title={Towards Conversational Diagnostic AI}, 
      author={Tao Tu and Anil Palepu and Mike Schaekermann and Khaled Saab and Jan Freyberg and Ryutaro Tanno and Amy Wang and Brenna Li and Mohamed Amin and Nenad Tomasev and Shekoofeh Azizi and Karan Singhal and Yong Cheng and Le Hou and Albert Webson and Kavita Kulkarni and S Sara Mahdavi and Christopher Semturs and Juraj Gottweis and Joelle Barral and Katherine Chou and Greg S Corrado and Yossi Matias and Alan Karthikesalingam and Vivek Natarajan},
      year={2024},
      eprint={2401.05654},
      archivePrefix={arXiv},
      primaryClass={cs.AI},
      url={https://arxiv.org/abs/2401.05654}, 
}

@inproceedings{chu2023multi,
author = {Chu, Zhendong and Wang, Nan and Wang, Hongning},
title = {Multi-objective Intrinsic Reward Learning for Conversational Recommender Systems},
year = {2023},
publisher = {Curran Associates Inc.},
address = {Red Hook, NY, USA},
booktitle = {Proceedings of the 37th International Conference on Neural Information Processing Systems},
articleno = {777},
numpages = {11},
location = {New Orleans, LA, USA},
series = {NIPS '23}
}

@misc{eisenstein2025dont,
      title={Don't lie to your friends: Learning what you know from collaborative self-play}, 
      author={Jacob Eisenstein and Reza Aghajani and Adam Fisch and Dheeru Dua and Fantine Huot and Mirella Lapata and Vicky Zayats and Jonathan Berant},
      year={2025},
      eprint={2503.14481},
      archivePrefix={arXiv},
      primaryClass={cs.LG},
      url={https://arxiv.org/abs/2503.14481}, 
}

@misc{fang2024multiagent,
      title={A Multi-Agent Conversational Recommender System}, 
      author={Jiabao Fang and Shen Gao and Pengjie Ren and Xiuying Chen and Suzan Verberne and Zhaochun Ren},
      year={2024},
      eprint={2402.01135},
      archivePrefix={arXiv},
      primaryClass={cs.IR},
      url={https://arxiv.org/abs/2402.01135}, 
}

@misc{surana2025reviews,
      title={From Reviews to Dialogues: Active Synthesis for Zero-Shot LLM-based Conversational Recommender System}, 
      author={Rohan Surana and Junda Wu and Zhouhang Xie and Yu Xia and Harald Steck and Dawen Liang and Nathan Kallus and Julian McAuley},
      year={2025},
      eprint={2504.15476},
      archivePrefix={arXiv},
      primaryClass={cs.IR},
      url={https://arxiv.org/abs/2504.15476}, 
}

@inproceedings{liang-etal-2024-llm,
    title = "{LLM}-{REDIAL}: A Large-Scale Dataset for Conversational Recommender Systems Created from User Behaviors with {LLM}s",
    author = "Liang, Tingting  and
      Jin, Chenxin  and
      Wang, Lingzhi  and
      Fan, Wenqi  and
      Xia, Congying  and
      Chen, Kai  and
      Yin, Yuyu",
    editor = "Ku, Lun-Wei  and
      Martins, Andre  and
      Srikumar, Vivek",
    booktitle = "Findings of the Association for Computational Linguistics: ACL 2024",
    month = aug,
    year = "2024",
    address = "Bangkok, Thailand",
    publisher = "Association for Computational Linguistics",
    url = "https://aclanthology.org/2024.findings-acl.529/",
    doi = "10.18653/v1/2024.findings-acl.529",
    pages = "8926--8939"
}

@inproceedings{kook-etal-2025-empowering,
    title = "Empowering Retrieval-based Conversational Recommendation with Contrasting User Preferences",
    author = "Kook, Heejin  and
      Kim, Junyoung  and
      Park, Seongmin  and
      Lee, Jongwuk",
    editor = "Chiruzzo, Luis  and
      Ritter, Alan  and
      Wang, Lu",
    booktitle = "Proceedings of the 2025 Conference of the Nations of the Americas Chapter of the Association for Computational Linguistics: Human Language Technologies (Volume 1: Long Papers)",
    month = apr,
    year = "2025",
    address = "Albuquerque, New Mexico",
    publisher = "Association for Computational Linguistics",
    url = "https://aclanthology.org/2025.naacl-long.392/",
    pages = "7692--7707",
    ISBN = "979-8-89176-189-6"
}

@inproceedings{joko2024laps,
author = {Joko, Hideaki and Chatterjee, Shubham and Ramsay, Andrew and de Vries, Arjen P. and Dalton, Jeff and Hasibi, Faegheh},
title = {Doing Personal LAPS: LLM-Augmented Dialogue Construction for Personalized Multi-Session Conversational Search},
year = {2024},
isbn = {9798400704314},
publisher = {Association for Computing Machinery},
address = {New York, NY, USA},
url = {https://doi.org/10.1145/3626772.3657815},
doi = {10.1145/3626772.3657815},
booktitle = {Proceedings of the 47th International ACM SIGIR Conference on Research and Development in Information Retrieval},
pages = {796–806},
numpages = {11},
location = {Washington DC, USA},
series = {SIGIR '24}
}

@misc{grattafiori2024llama3herdmodels,
      title={The Llama 3 Herd of Models}, 
      author={Grattafiori, Aaron and Dubey, Abhimanyu and Jauhri, Abhinav and Pandey, Abhinav and Kadian, Abhishek and Al-Dahle, Ahmad and Letman, Aiesha and Mathur, Akhil and Schelten, Alan and Vaughan, Alex and others},
      year={2024},
      eprint={2407.21783},
      archivePrefix={arXiv},
      primaryClass={cs.AI},
      url={https://arxiv.org/abs/2407.21783}, 
}

@misc{qwen3technicalreport,
      title={Qwen3 Technical Report}, 
      author={Qwen Team},
      year={2025},
      eprint={2505.09388},
      archivePrefix={arXiv},
      primaryClass={cs.CL},
      url={https://arxiv.org/abs/2505.09388}, 
}

@inproceedings{bert-score,
  title={BERTScore: Evaluating Text Generation with BERT},
  author={Tianyi Zhang* and Varsha Kishore* and Felix Wu* and Kilian Q. Weinberger and Yoav Artzi},
  booktitle={International Conference on Learning Representations},
  year={2020},
  url={https://openreview.net/forum?id=SkeHuCVFDr}
}

@article{harper2015movielens,
  author = {Harper, F. Maxwell and Konstan, Joseph A.},
  title = {The MovieLens Datasets: History and Context},
  year = {2015},
  journal = {ACM Trans. Interact. Intell. Syst.},
  volume = {5},
  number = {4},
  doi = {10.1145/2827872},
  url = {https://doi.org/10.1145/2827872}
}

@inproceedings{wang2022unicrs,
  title={Towards unified conversational recommender systems via knowledge-enhanced prompt learning},
  author={Wang, Xiaolei and Zhou, Kun and Wen, Ji-Rong and Zhao, Wayne Xin},
  booktitle={Proceedings of the 28th ACM SIGKDD conference on knowledge discovery and data mining},
  pages={1929--1937},
  year={2022}
}

@inproceedings{jiao2019tinybert,
  title = {TinyBERT: Distilling BERT for Natural Language Understanding},
  author = {Jiao, Xiaoqi and Yin, Yichun and Shang, Lifeng and Jiang, Xin and Chen, Xiao and Li, Linlin and Wang, Fang and Liu, Qun},
  booktitle = {Findings of the Association for Computational Linguistics: EMNLP 2020},
  year = {2020},
  publisher = {Association for Computational Linguistics},
  doi = {10.18653/v1/2020.findings-emnlp.372},
  url = {https://doi.org/10.18653/v1/2020.findings-emnlp.372}
}

@inproceedings{rendleBPR2009,
  title = {BPR: Bayesian Personalized Ranking from Implicit Feedback},
  author = {Rendle, Steffen and Freudenthaler, Christoph and Gantner, Zeno and Schmidt-Thieme, Lars},
  booktitle = {Proceedings of the 25th Conference on Uncertainty in Artificial Intelligence (UAI)},
  year = {2009},
  url = {https://www.auai.org/uai2009/papers/UAI2009_0139_48141db02b9f0b02bc7158819ebfa2c7.pdf}
}
\end{document}